\documentclass[sigconf, nonacm]{acmart}

\AtBeginDocument{%
  }

\copyrightyear{2026}
\acmYear{2026}
\setcopyright{cc}
\setcctype{by}
\acmConference[KDD '26]{Proceedings of the 32nd ACM SIGKDD Conference on Knowledge Discovery and Data Mining V.2}{August 09--13, 2026}{Jeju Island, Republic of Korea}
\acmBooktitle{Proceedings of the 32nd ACM SIGKDD Conference on Knowledge Discovery and Data Mining V.2 (KDD '26), August 09--13, 2026, Jeju Island, Republic of Korea}
\acmDOI{10.1145/3770855.3817553}
\acmISBN{979-8-4007-2259-2/2026/08}

\usepackage[T1]{fontenc}

\usepackage[utf8]{inputenc}

\usepackage{graphicx}

\usepackage{booktabs}
\usepackage{multirow}
\usepackage{amsmath}

\usepackage{csquotes}

\usepackage{caption}
\usepackage{subcaption}

\usepackage{threeparttable}

\usepackage{xurl}

\usepackage{doi}

\makeatletter
\renewcommand{\sectionautorefname}{\S\@gobble}
\renewcommand{\subsectionautorefname}{\S\@gobble}
\renewcommand{\subsubsectionautorefname}{\S\@gobble}

\makeatother

\usepackage{acro}[=v2]

\acsetup{display-foreign = false}
\acsetup{foreign-format = {\scshape} }
\ExplSyntaxOn
\prop_new:N \l__acro_foreign_format_prop
\ExplSyntaxOff

\DeclareAcronym{ai}{
    short = {AI},
    long = {Artificial Intelligence}
}

\DeclareAcronym{nlp}{
    short = {NLP},
    long = {Natural Language Processing}
}

\DeclareAcronym{ner}{
    short = {NER},
    long = {Named Entity Recognition}
}

\DeclareAcronym{re}{
    short = {RE},
    long = {Relation Extraction}
}

\DeclareAcronym{qa}{
    short = {QA},
    long = {Question Answering}
}

\DeclareAcronym{rag}{
    short = {RAG},
    long = {Retrieval-Augmented Generation}
}

\DeclareAcronym{llm}{
    short = {LLM},
    long = {Large Language Model}
}

\DeclareAcronym{gpt}{
    short = {GPT},
    long = {Generative Pre-trained Transformer}
}

\DeclareAcronym{sota}{
    short = {SOTA},
    long = {State-of-the-Art}
}

\DeclareAcronym{mlm}{
    short = {MLM},
    long = {Masked Language Model}
}

\DeclareAcronym{dlt}{
    short = {DLT},
    long = {Distributed Ledger Technology},
    short-plural = {s},
    long-plural-form = {Distributed Ledger Technologies}
}

\DeclareAcronym{p2p}{
    short = {P2P},
    long = {Peer-to-Peer},
}

\DeclareAcronym{dag}{
    short = {DAG},
    long = {Directed Acyclic Graph}
}

\DeclareAcronym{pos}{
    short = {PoS},
    long = {Proof of Stake}
}

\DeclareAcronym{pow}{
    short = {PoW},
    long = {Proof of Work}
}

\DeclareAcronym{evm}{
    short = {EVM},
    long = {Ethereum Virtual Machine}
}

\DeclareAcronym{dex}{
    short = DEX,
    long = Decentralized Exchange
}

\DeclareAcronym{defi}{
    short = DeFi,
    long = Decentralized Finance
}

\DeclareAcronym{amm}{
    short = AMM,
    long = Automated Market Maker
}

\DeclareAcronym{tvl}{
    short = TVL,
    long = Total Value Locked
}

\DeclareAcronym{cex}{
    short = CEX,
    long = Centralized Exchange
}

\DeclareAcronym{nft}{
    short = NFT,
    long = Non-Fungible Token
}

\DeclareAcronym{dapp}{
    short = DApp,
    long = Decentralized Application
}

\DeclareAcronym{dao}{
    short = DAO,
    long = Decentralized Autonomous Organization
}

\DeclareAcronym{ico}{
    short = ICO,
    long = Initial Coin Offering
}

\DeclareAcronym{mev}{
    short = MEV,
    long = Maximal Extractable Value
}

\DeclareAcronym{uspto}{
    short = USPTO,
    long = United States Patent and Trademark Office
}

\DeclareAcronym{micar}{
    short = MiCAR,
    long = Markets in Crypto-Assets Regulation
}

\DeclareAcronym{gdpr}{
    short = GDPR,
    long = General Data Protection Regulation
}

\DeclareAcronym{tcs}{
    short = T\&Cs,
    long = Terms and Conditions
}

\DeclareAcronym{esg}{
    short = ESG,
    long = {Environmental, Social, and Governance}
}

\DeclareAcronym{us}{
    short = US,
    long = {United States}
}

\begin{document}

\title{DLT-Corpus: A Large-Scale Text Collection for the Distributed Ledger Technology Domain}

\author{Walter Hernandez Cruz$^{1, 3}$, 
        Peter Devine$^{2}$, 
        Nikhil Vadgama$^{1, 3}$, 
        Paolo Tasca$^{1, 3}$, 
        Jiahua Xu$^{1, 3}$
        }
\affiliation{%
  \institution{
  $^{1}$Centre for Blockchain Technologies, University College London \\
  }
  \country{}
  \institution{$^{2}$School of Informatics, University of Edinburgh}
  \institution{$^{3}$Exponential Science Foundation}
}

\renewcommand{\authors}{Walter Hernandez Cruz, Peter Devine, Nikhil Vadgama, Paolo Tasca, Jiahua Xu}

\renewcommand{\shortauthors}{Hernandez Cruz et al.}

\email{{ walter.hernandez.18, nikhil.vadgama, p.tasca, jiahua.xu }@ucl.ac.uk,
 pdevine2@ed.ac.uk
}

\begin{abstract}
We introduce \acs{dlt}-Corpus, the largest domain-specific text collection for \acf{dlt} research to date: 2.98 billion tokens from 22.12 million documents spanning scientific literature (37,440 publications), \ac{uspto} patents (49,023 filings), and social media (22 million posts). 
Existing \ac{nlp} resources for \ac{dlt} focus narrowly on cryptocurrency price prediction and smart contracts, leaving domain-specific language underexplored despite the sector's $\sim$\$3 trillion market capitalization and rapid technological evolution.

We demonstrate \acs{dlt}-Corpus' utility by analyzing patterns of technology emergence and market-innovation correlations. 
Findings reveal that technologies first appear in our scientific literature subset before reaching patents and social media, following traditional technology transfer patterns. While social media sentiment remains overwhelmingly bullish even during crypto winters, scientific and patent activity grows less tied to short-term sentiment,
tracking overall market expansion in a virtuous cycle in which research precedes and enables economic growth that, in turn, funds further innovation.

We release the \ac{dlt}-Corpus and companion artifacts: LedgerBERT (+23\% over BERT-base on \ac{dlt}-specific \ac{ner} task), a sentiment analysis dataset of 23,301 crypto news headlines and descriptions, tools, and code.
\end{abstract}



\maketitle

\section{Introduction}

The \ac{dlt} field lacks a comprehensive text corpus. While, at the time of this writing, the ecosystem has grown to 
$\sim$\$3 trillion in market capitalization \cite{Budish2025TrustBlockchains} in early 2025
and introduced new concepts such as stablecoins, \acp{dex}, and \acp{amm} \cite{HernandezCruz2025EvolutionLiterature}, \ac{nlp} research within the \ac{dlt} domain remains constrained by narrow, task-specific datasets that overlook substantial textual resources in scientific publications, patents, and technical documentation.

Current \ac{dlt} datasets focus primarily on cryptocurrency price prediction \cite{McNally2018PredictingLearning,Seroyizhko2022APosts,Gurgul2025DeepData}, trading \cite{McNally2018PredictingLearning, Li2024CryptoTrade:Trading,Luo2026ResistingReasoning,Ding2025DecomposeMarket}, and smart contracts \cite{Chen2023ConversionNLP,Yang2023AutomatedModels,Sun2025EthereumLearning,Kim2024EthereumDistilBERT}. These resources support specific \ac{nlp} downstream tasks, such as \ac{ner} \cite{HernandezCruz2025EvolutionLiterature}, \ac{qa} \cite{Sarkar2025CryptOpiQA:Cryptocurrency}, sentiment analysis \cite{Azmina2022XLNET-GRUMalay,Rasivisuth2024AnReturn}, but fail to capture the substantial textual resources available in scientific literature, patent filings, and technical documentation. This gap limits the development of practical \ac{nlp} applications: \ac{rag} systems that reduce \ac{llm} hallucinations \cite{Sarkar2025CryptOpiQA:Cryptocurrency}, patent landscape monitoring, protocol documentation analysis, and technology trend detection, to cite a few examples. Similarly, \cite{Belcak2025SmallAI,Xiao2025LIMI:Agency,Pecher2025ComparingPerformance,Allal2025SmolLM2:Model,Juan2024Fine-TunedClassification,Grangier2024NeedEarly,Li2023AreTasks} demonstrate that small language models trained on domain-specific corpora outperform general-purpose \acp{llm} on specialized tasks while remaining computationally efficient \cite{Belcak2025SmallAI,Xiao2025LIMI:Agency,Li2023AreTasks,Juan2024Fine-TunedClassification}, highlighting the need for comprehensive \ac{dlt} text resources.

We introduce \textbf{DLT-Corpus}\footnote{\url{https://huggingface.co/collections/ExponentialScience/dlt-corpus}}, the largest domain-specific text collection for \ac{dlt} research: 2.98 billion tokens from 22.12 million documents spanning open-access scientific literature\footnote{\url{https://huggingface.co/datasets/ExponentialScience/DLT-Scientific-Literature}}, \ac{uspto} patent filings\footnote{\url{https://huggingface.co/datasets/ExponentialScience/DLT-Patents}}, and Twitter/X posts\footnote{\url{https://huggingface.co/datasets/ExponentialScience/DLT-Tweets}} collected before 2023 platform restrictions \cite{Davidson2023Platform-controlledScience}. This corpus integrates technical specifications, economic mechanisms, community discourse, and governance frameworks, providing a foundation for \ac{nlp} systems tailored to the \ac{dlt} domain.

We demonstrate the corpus utility through two analyses. First, we track how technologies diffuse from scientific literature to patents to social media, finding that concepts such as stablecoins, \acp{amm}, and \acp{dex} consistently appear first in research before reaching commercial and consumer communities. Second, we examine correlations between market dynamics and innovation activity, finding that scientific publications lead market expansion by two years ($\rho = 0.95$, $p < 0.001$), while social media sentiment remains bullish (i.e., extremely optimistic) even during crypto winters.

In summary, our domain resource contribution comprises:

\begin{itemize}
    \item \textbf{DLT-Corpus} (domain resource)\footnote{\url{https://huggingface.co/collections/ExponentialScience/dlt-corpus}}: 2.98 billion tokens from 22.12 million documents (37,440 scientific publications, 49,023 patents, 22M social media posts) with rich metadata enabling cross-disciplinary research.
    \item \textbf{Innovation diffusion analysis}: Evidence that \acfp{dlt} follow traditional technology transfer patterns, with research preceding market expansion and creating a virtuous funding cycle.
    \item \textbf{Sentiment analysis dataset}\footnote{\url{https://huggingface.co/datasets/ExponentialScience/DLT-Sentiment-News}}: 23,301 cryptocurrency news headlines and brief descriptions with crowdsourced annotations from active community members, addressing the need for domain-specific labeled data.
    \item \textbf{LedgerBERT}\footnote{\url{https://huggingface.co/ExponentialScience/LedgerBERT}}: A domain-adapted language model achieving 23\% improvement over BERT-base on \ac{dlt}-specific \ac{ner} task, developed through continued pre-training of SciBERT \cite{Beltagy2019SciBERT:Text}.
\end{itemize}

\acs{dlt}-Corpus, sentiment analysis dataset, models, and code\footnote{\url{https://github.com/dlt-science/DLT-Corpus}} are publicly available to support reproducibility and future research.

\section{Background}
\acf{dlt} refers to decentralized systems for recording and synchronizing data across multiple \ac{p2p} nodes using cryptographic techniques and consensus mechanisms. While blockchain represents the most recognized \ac{dlt} architecture, the term \ac{dlt} encompasses diverse architectures, including Parachains \cite{Wood2016Polkadot:Framework}, Sidechain \cite{Back2014EnablingSidechains}, Holochain \cite{Harris-BraunHolochainConsensus}, and \acp{dag} \cite{Raikwar2024SoK:Protocols} (e.g., Hashgraph \cite{Baird2020TheLedgers}), to cite a few examples. 
 
Therefore, while the term \blockquote{blockchain} is generically used for any distributed ledger system, we distinguish in this study between \blockquote{blockchain} (the chain-based architecture introduced by Bitcoin \cite{Nakamoto2008Bitcoin:System}) and \blockquote{\ac{dlt}} (the broader category including blockchain and other architectures) \cite{HernandezCruz2025EvolutionLiterature}.

\section{Related Work}

\paragraph{Existing \ac{dlt} text resources are fragmented and narrow}
The \ac{dlt} domain combines technical specifications \cite{Baird2020TheLedgers,Buterin2014Ethereum:Platform.,Nakamoto2008Bitcoin:System}, economic mechanisms \cite{Lo2020AssetsTokenomics,Moncada2024BlockchainTokenomics,Biais2025TheStaking}, and social dynamics \cite{HernandezCruz2025EvolutionLiterature,Cong2025BlockchainsChina}, with rapid terminological evolution introducing concepts like \acp{amm}, \acp{dex}, stablecoins, \acp{nft}, and \ac{mev}. Yet existing datasets focus narrowly on cryptocurrency markets: price prediction \cite{McNally2018PredictingLearning,Seroyizhko2022APosts,Gurgul2025DeepData,Kraaijeveld2020ThePrices}, trading \cite{Li2024CryptoTrade:Trading,PawlickaMaule2021CryptocurrencyDiscourse}, fraud detection \cite{Fu2025textscPerseus:Schemes,Li2025Knowledge-GroundedLMs}, and sentiment analysis \cite{Azmina2022XLNET-GRUMalay,Rasivisuth2024AnReturn,Guegan2021DoesPredictability}. These datasets primarily extract social media content from Twitter/X, Telegram, and Reddit \cite{Kraaijeveld2020ThePrices,Kang2024DecipheringTwitter,Seroyizhko2022APosts}, but despite some studies collecting millions of tweets, this data is rarely publicly available. Other work relies on transactional data \cite{Gai2023BlockchainModels,Sun2025EthereumLearning} and smart contracts \cite{Chen2023ConversionNLP,Yang2023AutomatedModels,Kim2024EthereumDistilBERT}, overlooking textual resources in scientific publications, patents, and technical documentation. General-purpose corpora such as RefinedWeb \cite{Penedo2023TheOnly}, CommonCrawl, and C4 \cite{Raffel2019ExploringTransformer} contain some cryptocurrency content but lack domain specificity. \ac{dlt}-Corpus addresses this gap by integrating scientific literature, \ac{uspto} patents, and Twitter/X data collected before 2023 API restrictions \cite{Davidson2023Platform-controlledScience}, enabling both cross-domain analysis and domain-specific \ac{nlp} applications.

\paragraph{Innovation diffusion in \ac{dlt} lacks integrated analysis.}
Prior work on \ac{dlt} domain analysis remains fragmented: \ac{ner} applied to scientific literature \cite{HernandezCruz2025EvolutionLiterature} or patents \cite{Yang2024NamedLearning}, news-based studies \cite{Perdana2021DistributedAhead}, taxonomies \cite{Tasca2019AClassification,Ballandies2022DecryptingEvaluation}, and systematic reviews \cite{Xu2019ABlockchain,Gorkhali2020Blockchain:Review}. Our work provides the first integrated analysis spanning scientific publications, patents, and community discourse (\autoref{sec:analysis}), revealing how innovation emerges and diffuses across research, commercial, and user communities.

\section{Datasets}

We introduce two datasets: 
\begin{itemize}
    \item \textbf{DLT-Corpus} (domain resource)\footnote{\url{https://huggingface.co/collections/ExponentialScience/dlt-corpus}}: 2.98 billion tokens from 22.12 million documents spanning scientific literature (37,440 publications), patents (49,023 filings), and social media (22M posts). This unstructured text corpus captures technical specifications, economic mechanisms, and community discourse across the \ac{dlt} domain.
    \item \textbf{Sentiment analysis dataset}\footnote{\url{https://huggingface.co/datasets/ExponentialScience/DLT-Sentiment-News}}: 23,301 cryptocurrency news headlines (and a brief description of them) with crowdsourced annotations from active community members, enabling sentiment analysis research.
\end{itemize}

We describe the construction of each dataset below. See \autoref{sec:datasets-documentation} and \autoref{sec:additional-datasets-documentation} for documentation and data dictionaries (\autoref{tab:fields-scientific}, \autoref{tab:fields-patents}, \autoref{tab:fields-tweets}, \autoref{tab:fields-sentiment}).

\subsection{\ac{dlt}-Corpus}

\begin{figure*}[tb!]
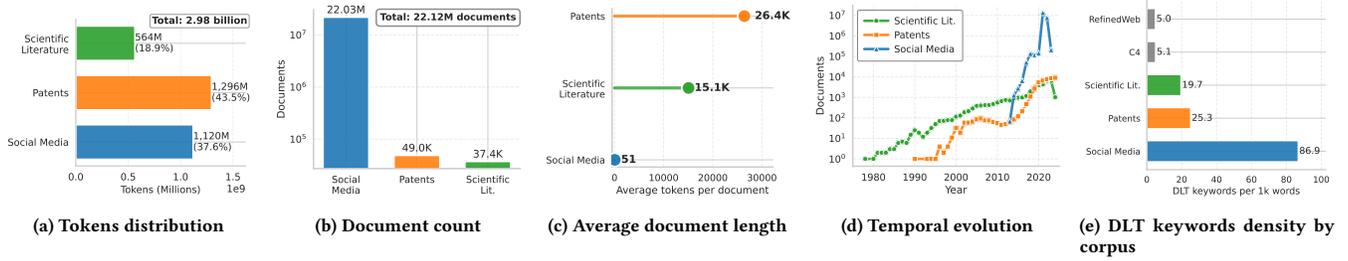

\centering
    \input{figures/token_distribution}
    \hfill
    \input{figures/document_count}
    \hfill
    \input{figures/avg_document_length}
    \hfill
    \input{figures/temporal_evolution}
    \hfill
    \input{figures/dlt_keywords_density}
    \captionsetup{font=small}
    \caption[Overview of \ac{dlt}-Corpus composition]{Overview of \ac{dlt}-Corpus composition}
    \label{fig:DLTCorpusOverview}
\end{figure*}

We construct the \ac{dlt}-Corpus by aggregating text from three complementary sources that capture different aspects of the \ac{dlt} ecosystem: academic and industry publications provide formal technical knowledge, patent filings reveal innovation trajectories, and social media reflect community discourse and market dynamics. This multi-source approach allows comprehensive coverage of technical terminology and evolving community language.

\paragraph{Legal compliance} We exclusively use open-access scientific literature and, whenever available, include the license for each publication in the metadata (\autoref{tab:fields-scientific}) to ensure redistribution rights for research and commercial applications. Patent data derives from public \ac{uspto} records. Social media content was collected before Twitter/X's 2023 \ac{tcs} changes that restricted API access on 18 May 2023\footnote{\url{https://x.com/en/tos/previous/version\_18}}, preserving a valuable snapshot of community discourse. Then, the \ac{dlt}-Corpus could facilitate academic research and industry use by mitigating legal barriers, which have become a persistent challenge in the \ac{nlp} domain \cite{Castilho2018AProcessing}, especially for industry applications with commercial benefits.

\paragraph{Rich metadata} Each subset includes structured metadata enabling research beyond language modeling: publication venues, authors, and references for scientific literature (\autoref{tab:fields-scientific}). Inventor, assignee, and filing information for patents (\autoref{tab:fields-patents}). Timestamps and sentiment labels (based on \autoref{subsec:sentiment-evaluation}) for social media (\autoref{tab:fields-tweets}). Therefore, the metadata included for each source of the \ac{dlt}-Corpus could support cross-disciplinary investigations spanning innovation diffusion, technology forecasting, collaborative network analysis, computational social science, and other areas of research reflecting the increasingly interdisciplinary nature of the \ac{dlt} field \cite{HernandezCruz2025EvolutionLiterature}. For industry practitioners, this metadata could enable practical applications such as patent landscape monitoring, R\&D trend detection, and competitor analysis.

\paragraph{Corpus statistics.} \autoref{fig:DLTCorpusOverview} summarizes the corpus: 2.98 billion tokens across 22.12 million documents. Document lengths vary by source: social media averages 51 tokens, scientific literature 15.1k tokens, and patents 26.4k tokens. Token distribution: patents 43.5\%, social media 37.6\%, scientific literature 18.9\% (\autoref{fig:TokenDistribution}).

\subsubsection{Scientific literature} \label{subsec:industry_and_academy_dataset}

\paragraph{Collection} We retrieved PDFs and metadata (authors, title, year, venue, references, licensing)\footnote{Copyright licensing information included when available via Semantic Scholar.} from Semantic Scholar\footnote{\url{https://www.semanticscholar.org/}} using domain-specific queries (e.g., \blockquote{\acl{dlt}}, \blockquote{Blockchain}, \blockquote{Hashgraph}, \blockquote{DAG}, \blockquote{Consensus Mechanisms}, \blockquote{Distributed Computing}, \blockquote{Distributed Storage}, \blockquote{Distributed Databases}), producing 144,843 initial documents.

\paragraph{Processing} We parsed PDFs to Markdown using PyMuPDF4LLM,\footnote{\url{https://pypi.org/project/pymupdf4llm/}} filtered for English-only content using FastText \cite{Joulin2017BagClassification}, and removed outliers that were either too short ($<$ 500 tokens) or too long ($>$ 40k tokens), retaining 112,911 documents.

\paragraph{Domain filtering} To ensure relevance, we fine-tuned BERT-base-cased\footnote{\url{https://huggingface.co/google-bert/bert-base-cased}} on the \ac{ner} dataset from \cite{HernandezCruz2025EvolutionLiterature} and predicted domain-specific entities in each document. We filtered documents based on prediction quality by calculating: (1) the number of predicted entities per document, (2) the maximum prediction score, and (3) the median prediction score across all entities. Documents were retained if they had a maximum prediction score above $0.995$ or a median prediction score at or above the subset-wide median for the scientific literature, ensuring high confidence in domain-specific content while maintaining reasonable coverage. This filtering step, combined with duplicate removal, reduced the dataset to 38,010 documents.  

Then, manual review removed 570 marginally relevant papers, obtaining 37,440 publications. The removed papers were false positives, retrieved because terms such as \blockquote{distributed}, \blockquote{consensus}, \blockquote{protocol}, and \blockquote{network} appear in non-\ac{dlt} contexts. Manual inspection revealed these papers clustered around biomedical domains: Alzheimer's disease research (neurofibrillary tangles, tau proteins, amyloid, dementia), infectious disease epidemiology (COVID-19, tuberculosis), oncology, and healthcare systems because they share terminology with \ac{dlt} but focused on \blockquote{distributed} biological processes, clinical \blockquote{consensus protocols}, or \blockquote{decentralized} healthcare delivery rather than \acf{dlt}.

The scientific literature subset\footnote{\url{https://huggingface.co/datasets/ExponentialScience/DLT-Scientific-Literature}} spans 1978--mid 2025 and represents, to our knowledge, the first large-scale scientific corpus for the \ac{dlt} domain.

\subsubsection{Patents} \label{subsec:patents_dataset}

We collected patent text and metadata (e.g., patent number, title, publication date, inventor, assignee, applicant, database) from \ac{uspto}'s US-PGPUB and USPAT databases using search terms \blockquote{\acl{dlt}} and \blockquote{blockchain} in titles, abstracts, and full text. We focus on US patents because \ac{uspto}'s terms state that patent text is \blockquote{typically not subject to copyright restrictions}\footnote{\url{https://www.uspto.gov/terms-use-uspto-websites}}, facilitating \ac{nlp} research and commercial use.

The patents subset\footnote{\url{https://huggingface.co/datasets/ExponentialScience/DLT-Patents}} contains 49,023 documents spanning 1990--mid 2025 and represents, to our knowledge, the first compiled patent dataset for the \ac{dlt} domain.

\subsubsection{Social media} \label{subsec:social_media_dataset}

We aggregated Twitter/X data from academic \cite{jahanbin2023Database2021-2023.,Garg2021CrypTop12:Prices,Nizzoli2020ChartingManipulation} and industry sources,\footnote{\url{https://www.kaggle.com/datasets/leoth9/crypto-tweets}}$^{,}$\footnote{\url{https://www.kaggle.com/datasets/kaushiksuresh147/bitcoin-tweets}}$^{,}$\footnote{\url{https://www.kaggle.com/datasets/tleonel/crypto-tweets-80k-in-eng-aug-2022}}$^{,}$\footnote{\url{https://www.kaggle.com/datasets/rezasemyari/crypto-sentiment-tweets}}$^{,}$\footnote{\url{https://www.kaggle.com/datasets/hiraddolatzadeh/bitcoin-tweets-2021-2022/data}} all collected before Twitter/X's 2023 API restrictions that affected researcher access \cite{Davidson2023Platform-controlledScience}. Initial aggregation produced 28,775,339 posts. After removing empty posts (407) and deduplication (25,417,108 unique), we filtered for English using Lingua,\footnote{\url{https://github.com/pemistahl/lingua-py}}, resulting in 22,033,090 posts. Each post includes timestamp and sentiment labels (bullish, bearish, neutral) from LedgerBERT (\autoref{subsec:ledgerBERT}).

The social media subset\footnote{\url{https://huggingface.co/datasets/ExponentialScience/DLT-Tweets}} spans 2013--mid 2023 and represents, to our knowledge, the largest publicly available snapshot of \ac{dlt} community discourse on Twitter/X.

\subsubsection{Corpus quality assessment} \label{subsec:quality_assessment}

To quantify domain specificity and quality, we compare \ac{dlt}-Corpus vocabulary distributions against two general-purpose corpora: RefinedWeb \cite{Penedo2023TheOnly} (600B tokens) and C4 \cite{Raffel2019ExploringTransformer} (156B tokens). We curated 361 keywords from the \ac{dlt} taxonomy of \cite{HernandezCruz2025EvolutionLiterature} and \cite{Tasca2019AClassification}, covering consensus mechanisms, cryptographic primitives, smart contracts, token standards, \ac{defi} concepts, and major platforms. We sampled approximately 60M tokens from each corpus and computed keyword density (occurrences per 1,000 words), document coverage (percentage of documents containing at least one keyword), and Jensen-Shannon divergence to measure distributional differences \cite{Lu2020DivergingTasks}.

\begin{table}[t]
\centering
\captionsetup{font=small}
\caption{Comparison of \ac{dlt}-Corpus with RefinedWeb \cite{Penedo2023TheOnly} and C4 \cite{Raffel2019ExploringTransformer}}
\label{tab:vocab_benchmark}
\footnotesize
\begin{tabular}{llccc}
\toprule
\textbf{Corpus} & \textbf{Type} & \textbf{Density} & \textbf{Doc. Cov.} & \textbf{JS Div.} \\
\midrule
Social Media & Domain & 86.92 & 96.3\% & 0.44 \\
Patents & Domain & 25.27 & 99.8\% & 0.45 \\
Scientific Lit. & Domain & 19.68 & 100.0\% & 0.39 \\
\midrule
\textit{DLT-Corpus avg.} & \textit{Domain} & \textit{43.96} & \textit{98.7\%} & \textit{0.43} \\
\midrule
RefinedWeb & General & 4.96 & 55.6\% & 0.10 \\
C4 & General & 5.14 & 51.9\% & 0.10 \\
\bottomrule
\end{tabular}

\begin{tablenotes}
    \footnotesize
    \item \textit{-Keyword density:} vocabulary analysis comparing \ac{dlt}-Corpus subsets against general-purpose web corpora using 361 domain keywords. Keyword density measures occurrences per 1,000 words.
    \item \textit{-Document coverage:} indicates the percentage of documents containing at least one domain keyword
    \item \textit{-Jensen-Shannon (JS) divergence:} measures vocabulary distribution difference versus general corpora (higher values indicate greater difference); for general corpora, the value shows the baseline divergence between RefinedWeb and C4.
\end{tablenotes}

\end{table}

\ac{dlt}-Corpus exhibits 8.7 times higher keyword density than general corpora (see \autoref{tab:vocab_benchmark}): 43.96 keywords per 1,000 words (averaging across subsets) versus 5.05 in RefinedWeb and C4 (see also \autoref{fig:DLTKeywordsDensity}). Document coverage reaches 98.7\% in \ac{dlt}-Corpus compared to 53.8\% in general corpora. Jensen-Shannon divergence between \ac{dlt}-Corpus and general corpora ranges from 0.39 to 0.45, while divergence between RefinedWeb and C4 is only 0.10. This difference confirms that \ac{dlt}-Corpus has fundamentally distinct vocabulary distributions from general web text.

Higher keyword density matters for language model learning because language models rely on word frequency during training to acquire vocabulary \cite{Chang2022WordModels}. Domain-adaptive pretraining on corpora with concentrated terminology exposure leads to performance gains on downstream tasks \cite{Gururangan2020DontTasks}, as models encounter domain-specific terms more frequently and learn their contextual usage patterns \cite{Chang2022WordModels}. The higher density of 8.7 in \ac{dlt}-Corpus, compared to general web text, provides substantially more learning signal per token for \ac{dlt} terminology.

\subsection{Sentiment analysis} \label{subsec:sentiment_analysis_dataset}

To support sentiment analysis in the \ac{dlt} domain, we constructed a labeled dataset\footnote{\url{https://huggingface.co/datasets/ExponentialScience/DLT-Sentiment-News}} from CryptoPanic,\footnote{Data collected via CryptoPanic's free API, March--May 2025. \ac{tcs} at collection time contained no restrictions on academic research.} a cryptocurrency news platform where active community members vote on articles.

\paragraph{Annotation.} Users vote on news headlines and brief descriptions of them across three dimensions: market direction (bullish/bearish), content characteristics (important/lol\footnote{\blockquote{lol} (Laughing Out Loud) indicates humorous headlines.}), and engagement quality (liked/disliked). We normalize vote percentages by total engagement, filter by the median minimum votes, and use the 25th/75th percentiles as classification boundaries: below the 25th = negative, above the 75th = positive, between = neutral. This percentile-based approach mitigates popularity bias from raw counts.

\paragraph{Crowdsourcing advantages.} Our approach leverages collective intelligence \cite{Snow2008CheapTasks} from domain experts (active crypto users), avoiding the pitfalls of an \ac{llm}-based annotation approach, which, if we had followed it, could have introduced systematic biases and statistical manipulation \cite{Baumann2025LargeAnnotation}.

\paragraph{Statistics.} The dataset contains 23,301 examples (1.85M tokens, 79.51 tokens/example average) spanning 2021--mid 2025.

\section{Companion language model} \label{subsec:ledgerBERT}

To demonstrate the practical utility of \ac{dlt}-Corpus for language model development, we train LedgerBERT\footnote{\url{https://huggingface.co/ExponentialScience/LedgerBERT}}, a companion domain-adapted encoder for \ac{dlt}-specific \ac{nlp} tasks. We evaluate LedgerBERT on two tasks: in-domain \ac{ner} (where improvement validates corpus quality) and out-of-domain sentiment analysis (where maintained performance validates generalization).

\paragraph{Training} We use continued pre-training rather than training from scratch, following evidence that domain adaptation of existing models outperforms full pre-training \cite{Xie2024EfficientModels,Gururangan2020DontTasks}. We initialize from SciBERT \cite{Beltagy2019SciBERT:Text}, which captures multidisciplinary scientific content and likely includes some \ac{dlt}-related material, making it a stronger starting point than general-purpose BERT.

\paragraph{Hyperparameters.} We experimented with different hyperparameter configurations and selected final values based on model convergence and validation loss. We train for 3 epochs with learning rate $5 \times 10^{-5}$ (linear decay), \ac{mlm} probability 0.15, warmup ratio 0.10, batch size 12, sequence length 512, weight decay 0.01, and Stable AdamW optimizer \cite{Wortsman2023StableModels} with bfloat16 precision.

\paragraph{Compute} Training on one NVIDIA H100 GPU required approximately 
68.7 GPU-hours.

\subsection{Corpus quality evaluation: in-domain \ac{ner}}
\label{subsec:ner-evaluation}

\begin{table}[t]
\centering

\captionsetup{size=small}
\caption[Performance on in-domain \ac{dlt} entity recognition from scientific literature]{Performance on in-domain \ac{dlt} entity recognition from scientific literature. \ac{ner} scores represent F1-average across 5-fold cross-validation with strict entity-level matching (exact boundary and type agreement required).}
\label{tab:ner_results}

\footnotesize
\begin{tabular}{lcc}
\toprule
\textbf{Model} & \textbf{Parameters} & \textbf{NER F1} \\
\midrule
\multicolumn{3}{l}{\textit{General Baselines}} \\
BERT-base \cite{Devlin2019BERT:Understanding} & 110M & 0.243 \\
MosaicBERT-base \cite{Portes2023MosaicBERT:Pretraining} & 109M & 0.076 \\
ModernBERT-base \cite{Warner2025SmarterInference} & 149M & 0.209 \\
\midrule
\multicolumn{3}{l}{\textit{Domain-Adapted Baselines}} \\
SciBERT \cite{Beltagy2019SciBERT:Text} & 110M & 0.289 \\
BioClinical ModernBERT \cite{Sounack2025BioClinicalNLP} & 150M & 0.192 \\
\midrule
\multicolumn{3}{l}{\textit{LedgerBERT (This work)}} \\
LedgerBERT & 110M & \textbf{0.299} \\
\quad \textit{Improvement over SciBERT} & & \textit{+3.5\%} \\
\quad \textit{Improvement over BERT-base} & & \textit{+23.05\%} \\
\bottomrule
\end{tabular}
\end{table}

\ac{ner} serves as the primary evaluation of corpus quality because performance directly reflects how well the model learned domain-specific terminology. We use the \ac{dlt}-focused \ac{ner} dataset from \cite{HernandezCruz2025EvolutionLiterature}, targeting entities such as consensus mechanisms (\ac{pos}, \ac{pow}), platforms (Ethereum, Hedera), and technical concepts (Merkle tree, private key). This dataset derives from scientific literature, matching our corpus composition.

\paragraph{Fine-tuning.} We experimented with learning rates and number of epochs, selecting final hyperparameters based on convergence behavior and cross-validation performance. We fine-tune for 20 epochs with learning rate $1 \times 10^{-5}$, 500 warmup steps, batch size 16 (gradient accumulation 2 for effective batch 32), and 5-fold cross-validation grouped by paper to prevent data leakage.

\paragraph{Results.} \autoref{tab:ner_results} reports F1 scores using strict entity-level matching (exact boundary and type required). LedgerBERT achieves \textbf{0.299 F1}, improving over SciBERT (0.289 F1) by 3.5\% relative and over BERT-base (0.243 F1) by 23\%. The progression of BERT-base $\rightarrow$ SciBERT $\rightarrow$ LedgerBERT demonstrates the cumulative value of domain-specific pre-training: general scientific knowledge (SciBERT) provides a foundation, and \ac{dlt}-Corpus adds specialized terminology.

\subsection{Generalization test: out-of-domain sentiment analysis}
\label{subsec:sentiment-evaluation}

\begin{table}[t]
\centering

\captionsetup{font=small}
\caption{Performance on out-of-domain sentiment analysis from cryptocurrency news articles titles and brief descriptions. 
}
\label{tab:sentiment_results}

\footnotesize
\begin{tabular}{lccc}
\toprule
\textbf{Model} & \textbf{Market} & \textbf{Content} & \textbf{Engagement} \\
 & \textbf{Direction} & \textbf{Characteristics} & \textbf{Quality} \\
\midrule
BERT-base & 0.577 & 0.514 & 0.539 \\
DistilBERT-base & 0.570 & 0.508 & 0.544 \\
SciBERT & 0.591 & 0.514 & 0.532 \\
\midrule
LedgerBERT & 0.590 & 0.502 & 0.537 \\
\quad \textit{vs. SciBERT} & \textit{-0.2\%} & \textit{-2.3\%} & \textit{+0.9\%} \\
\bottomrule
\end{tabular}

\begin{tablenotes}
    \footnotesize
    \item \textit{Note:} News articles are absent from \ac{dlt}-Corpus due to copyright restrictions on collecting, using, and redistributing journalistic content. Results demonstrate preserved general capabilities despite domain-specific training.
\end{tablenotes}

\end{table}

To verify that domain-specific training does not degrade general capabilities, we evaluate on sentiment analysis of cryptocurrency news, which is a task representing \textit{out-of-domain} generalization because news articles are absent from \ac{dlt}-Corpus due to copyright restrictions.\footnote{For example, see ongoing copyright lawsuits by news organizations against OpenAI \cite{Brittain2025JudgeReuters,Pope2024NYTReview} and Microsoft \cite{Pope2024NYTReview} over alleged copyright violations of news articles, which motivated this exclusion.}

\paragraph{Fine-tuning.} We selected hyperparameters through experimentation, monitoring convergence on the validation set. We fine-tune LedgerBERT\footnote{\url{https://huggingface.co/ExponentialScience/LedgerBERT-Market-Sentiment}} for 3 epochs with learning rate $2 \times 10^{-5}$, 500 warmup steps, batch size 8, and 90/10 train/test split.

\paragraph{Results.} \autoref{tab:sentiment_results} shows LedgerBERT performs comparably to SciBERT across all sentiment dimensions (within 0.2\% on market direction, the primary metric). This result is important: domain-specific training preserved general language understanding despite the corpus emphasizing scientific literature, patents, and social media posts rather than sentiment-bearing news.

\paragraph{Interpretation.} The NER improvement (23\% over BERT-base) combined with maintained sentiment performance demonstrates that \ac{dlt}-Corpus enables domain specialization without catastrophic forgetting, which means the model gains \ac{dlt}-specific knowledge while retaining general capabilities for out-of-domain tasks.

\section{Dataset documentation} \label{sec:datasets-documentation}

We provide standardized documentation following the Datasheet for Datasets framework \cite{Gebru2021DatasheetsDatasets} and align with the FAIR Guiding Principles \cite{Wilkinson2016TheStewardship} to ensure that our datasets are Findable, Accessible, Interoperable, and Reusable (FAIR) for human researchers and computational agents. See \autoref{subsec:sentiment-analysis-documentation} for documentation of the sentiment analysis dataset and \autoref{sec:additional-dlt-corpus-documentation} for additional documentation of \ac{dlt}-Corpus.

\subsection{\acs{dlt}-Corpus}

\subsection*{Motivation}

\textbf{Purpose:} \ac{dlt}-Corpus\footnote{\url{https://huggingface.co/collections/ExponentialScience/dlt-corpus}} was created to address the lack of 
large-scale, domain-specific text corpora for \ac{nlp} and other type of research in the \acf{dlt} field.

\textbf{Creators:} Walter Hernandez Cruz, Peter Devine, Nikhil Vadgama, Paolo Tasca, Jiahua Xu

\subsection*{Composition}

\textbf{Content:} 2.98 billion tokens across three subsets:
\begin{itemize}
    \item Scientific literature: 37,440 documents, 564M tokens
    \item Patents: 49,023 documents, 1,296M tokens  
    \item Social media: 22.03M documents, 1,120M tokens
\end{itemize}

\textbf{Temporal coverage:}
\begin{itemize}
    \item Scientific: 1978-2025
    \item Patents: 1990-2025
    \item Social media: 2013-mid 2023
\end{itemize}

\textbf{Language:} English

\textbf{Missing data:} Social media posts after 2023 due to platform access restrictions.

\textbf{Confidentiality:} No private or confidential data. All sources are publicly accessible. Social media usernames removed to protect privacy (see \autoref{sec:ethics}).

\textbf{Data fields}: 

\textit{Scientific literature:} \autoref{tab:fields-scientific} describes the fields in the scientific literature subset.
\begin{table}[t]

\footnotesize
\centering
\captionsetup{size=small}
\caption{Fields in Scientific Literature dataset}
\label{tab:fields-scientific}

\begin{tabular}{ll}
\toprule
\textbf{Field} & \textbf{Description} \\
\midrule
\texttt{paperId} & Semantic Scholar identifier \\
\texttt{title} & Publication title \\
\texttt{abstract} & Publication abstract \\
\texttt{text} & Full text in Markdown format \\
\texttt{url} & URL to the source document \\
\texttt{year} & Publication year \\
\texttt{publicationDate} & Full publication date \\
\texttt{venue} & Venue name \\
\texttt{publicationVenue} & Publication venue (journal, conference, etc.) \\
\texttt{publicationTypes} & Type of publication (e.g., JournalArticle) \\
\texttt{authors} & Author list with metadata \\
\texttt{references} & Cited works \\
\texttt{fieldsOfStudy} & Academic disciplines \\
\texttt{s2FieldsOfStudy} & Semantic Scholar's field classifications \\
\texttt{isOpenAccess} & Open access flag \\
\texttt{openAccessPdf} & Link to open access PDF if available and licensing \\
\texttt{lang} & Language code \\
\texttt{lang\_conf} & Language detection confidence \\
\texttt{total\_tokens} & Total number of tokens \\
\bottomrule
\end{tabular}
\end{table}

\textit{Patents:} Table~\ref{tab:fields-patents} describes the fields in the patents subset.
\begin{table}[ht]

\footnotesize
\centering
\captionsetup{size=small}
\caption{Fields in Patents dataset}
\label{tab:fields-patents}

\begin{tabular}{ll}
\toprule
\textbf{Field} & \textbf{Description} \\
\midrule
\texttt{Patent Number} & USPTO patent number (e.g., US10123456B2) \\
\texttt{Document ID} & Unique document identifier \\
\texttt{Title} & Patent title \\
\texttt{text} & Full patent text \\
\texttt{Date Published} & Publication date \\
\texttt{Filing Date} & Application filing date \\
\texttt{Application Number} & USPTO application number \\
\texttt{Family ID} & Patent family identifier \\
\texttt{Inventor} & Inventor names \\
\texttt{Assignee} & Patent owner \\
\texttt{Applicant Name} & Applicant information \\
\texttt{Primary Examiner} & USPTO primary examiner \\
\texttt{Assistant Examiner} & USPTO assistant examiner \\
\texttt{CPCI} & Cooperative Patent Classification Invention codes \\
\texttt{CPCA} & Cooperative Patent Classification Additional codes \\
\texttt{OR} & Other references \\
\texttt{XREF} & Related patent references \\
\texttt{Notes} & Additional notes \\
\texttt{Notes/Tagged} & Tagged annotations \\
\texttt{Relevancy} & Relevancy score or classification \\
\texttt{Database} & Source database (USPGPUB or USPAT) \\
\texttt{total\_tokens} & Total number of tokens in the document \\
\bottomrule
\end{tabular}
\end{table}

\textit{Tweets:} \autoref{tab:fields-tweets} describes the fields in the social media subset.
\begin{table}[ht]

\footnotesize
\centering
\captionsetup{size=small}
\caption{Fields in Social Media dataset}
\label{tab:fields-tweets}

\begin{tabular}{ll}
\toprule
\textbf{Field} & \textbf{Description} \\
\midrule
\texttt{tweet} & Tweet text \\
\texttt{timestamp} & Post creation time \\
\texttt{year} & Post year \\
\texttt{language} & Detected language \\
\texttt{sentiment\_class} & Sentiment category \\
\texttt{sentiment\_label} & Sentiment label \\
\texttt{sentiment\_score} & Sentiment score \\
\texttt{confidence\_level} & Prediction confidence for sentiment label \\
\texttt{total\_tokens} & Token count \\
\bottomrule
\end{tabular}
\end{table}

\subsection*{Collection}

\textbf{Scientific literature\footnote{\url{https://huggingface.co/datasets/ExponentialScience/DLT-Scientific-Literature}}:} Collected from Semantic Scholar API using domain-specific queries, filtered for domain relevance using fine-tuned BERT model (\autoref{subsec:industry_and_academy_dataset}).

\textbf{Patents\footnote{\url{https://huggingface.co/datasets/ExponentialScience/DLT-Patents}}:} Retrieved from \ac{uspto} public databases (USPGPUB, USPAT) using keyword searches (\autoref{subsec:patents_dataset}).

\textbf{Social media\footnote{\url{https://huggingface.co/datasets/ExponentialScience/DLT-Tweets}}:} Aggregated from previously published academic datasets and publicly available industry sources, all collected before Twitter/X's 2023 API restrictions (\autoref{subsec:social_media_dataset}).

\subsection*{Preprocessing}

\textbf{Scientific literature:} PDF parsing to Markdown, language detection, length filtering, domain relevance filtering (see \autoref{subsec:industry_and_academy_dataset} for more details).

\textbf{Patents:} Text extraction, formatting standardization (e.g., fix encoding errors).

\textbf{Social media:} Username removal, duplicate detection, language filtering (see \autoref{subsec:social_media_dataset} for more details).

\section{Analysis} \label{sec:analysis}

We demonstrate the utility of \ac{dlt}-Corpus through two analyses: (1) correlations between market dynamics and document production, and (2) technology diffusion patterns across communities. These analyses serve as examples of research enabled by the corpus.

\begin{figure*}[t]
\centering
\begin{minipage}[t!]{0.25\textwidth}
    \centering
     \includegraphics[width=\linewidth]{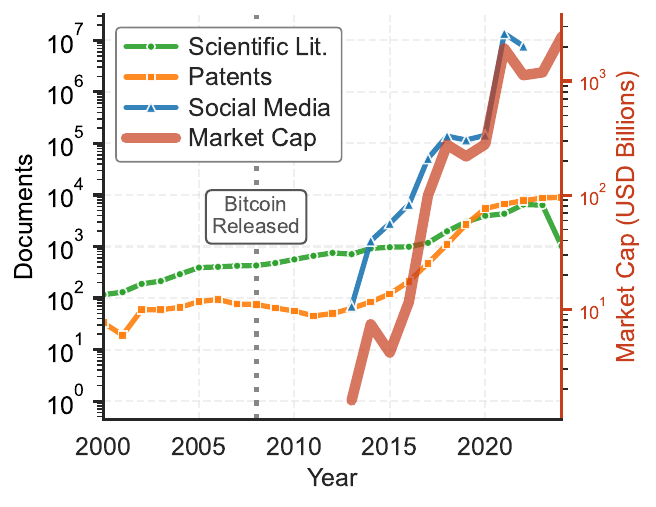}
     \captionsetup{font=small}
     \caption[Yearly growth of documents \& market capitalization]{Yearly growth of global cryptocurrency market capitalization\footnotemark and documents in the \ac{dlt}-Corpus.}
     \label{fig:TemporalMarketCap}
\end{minipage}
\hfill
\begin{minipage}[t!]{0.72\textwidth}
    \centering
    \input{figures/stablecoins}
    \hfill
    \input{figures/dex}
    \hfill
    \input{figures/amm}
    \captionsetup{font=small}
    \caption[Mentions per year for selected technologies in the \ac{dlt}-Corpus]{Mentions per year for selected technologies in the \ac{dlt}-Corpus. The y-axis represents the relative frequency of mentions normalized by the total volume of documents in each source per year.}
    \label{fig:TechnologiesMentionsPerYear}
\end{minipage}%
\end{figure*}

\paragraph{Market-document correlations.} \autoref{fig:TemporalMarketCap} shows document growth across all corpus subsets correlating with cryptocurrency market capitalization. We quantify these relationships using Spearman's rank correlation.

\begin{table}[t]
\centering
\captionsetup{size=small}
\caption[Correlation between cryptocurrencies' market capitalization and document volumes]{Correlation between cryptocurrencies' market capitalization and document volumes (\autoref{fig:TemporalMarketCap}).}
\label{tab:market-correlations}

\footnotesize
\begin{tabular}{lccc}
\toprule
\textbf{Document Type} & \textbf{Period} & \textbf{Spearman's $\rho$} & \textbf{$p$-value} \\
\midrule
Scientific literature & 2013-2024 & 0.76 & $<$0.004 \\
Patents & 2013-2024 & 0.96 & $<$0.001 \\
Social media & 2013-2023 & 0.98 & $<$0.001 \\
\bottomrule
\end{tabular}
\end{table}

\begin{table}[t]
\centering
\captionsetup{font=small}
\caption[Lagged correlations (Spearman's $\rho$) between document volumes and market capitalization]{Lagged correlations (Spearman's $\rho$) between document volumes and market capitalization.}
\label{tab:lagged-correlations}

\footnotesize
\begin{tabular}{@{}clll@{}}
\toprule
\textbf{Lag (years)} & \textbf{Scientific} & \textbf{Social Media} & \textbf{Patents} \\
\midrule
\multicolumn{4}{@{}l}{\textit{Document volumes lead market}} \\
$-5$ & 0.86** & 0.77 & 0.86** \\
$-4$ & 0.90** & 0.82* & 0.90** \\
$-3$ & 0.93*** & 0.95*** & 0.93*** \\
$-2$ & 0.95*** & 0.88** & 0.95*** \\
$-1$ & 0.94*** & 0.90*** & 0.95*** \\
\midrule
\multicolumn{4}{@{}l}{\textit{Concurrent}} \\
$\phantom{-}0$ & 0.76** & 0.98*** & 0.97*** \\
\midrule
\multicolumn{4}{@{}l}{\textit{Market leads document volumes}} \\
$1$ & 0.79** & 0.92*** & 0.95*** \\
$2$ & 0.68* & 0.85** & 0.96*** \\
$3$ & 0.47 & 0.86** & 0.97*** \\
$4$ & 0.31 & 0.89** & 0.95*** \\
$5$ & 0.36 & 0.60 & 0.93*** \\
\bottomrule
\end{tabular}

\begin{tablenotes}
\centering
\footnotesize
    \item $^{*}p<0.05$, $^{**}p<0.01$, $^{***}p<0.001$
\end{tablenotes}

\end{table}

\footnotetext{Median market capitalization per year of the cryptocurrency market using aggregated
data from CoinGecko.}

\autoref{tab:market-correlations} reports Spearman's rank correlations between annual market capitalization and document volumes (2013--2023 for social media, 2013--2024 for patents and publications).\footnote{Systematic market data became available around 2013 \cite{CoinMarketCap2018CoinMarketCapsAnniversary} as cryptocurrencies developed sufficient liquidity.} All three document types show strong positive correlations: scientific literature ($\rho = 0.76$, $p < 0.004$), patents ($\rho = 0.96$, $p < 0.001$), and social media ($\rho = 0.98$, $p < 0.001$). 

\paragraph{Lagged correlations reveal temporal structure.} To test whether research drives market expansion or responds to it, we compute lagged correlations (\autoref{tab:lagged-correlations}). The scientific literature subset spans 1978--2025 (earliest: \cite{Rivest1978ACryptosystems}), enabling meaningful lag analysis.

Scientific publications show asymmetric temporal patterns: correlations remain strong when publications lead the market (negative lags), but decay rapidly when the market leads publications and lose significance beyond two years ($\rho = 0.47$, $p > 0.05$ at three years). This asymmetry indicates research precedes market expansion.

Social media exhibits broadly strong correlations across nearly all lags, peaking concurrently ($\rho = 0.98$, $p < 0.001$) and remaining significant from lag $-4$ to lag $4$, with non-significance only at the five-year extremes ($\rho = 0.77$ when social media leads, $\rho = 0.60$ when the market leads). Patents show symmetric patterns with peak concurrent correlation ($\rho = 0.97$, $p < 0.001$), significant whether patents lead ($\rho = 0.93$, $p < 0.001$ at three years) or lag ($\rho = 0.97$, $p < 0.001$ at three years). 

Additionally, the three corpus subsets reflect distinct communities with different incentives: researchers seeking to disseminate knowledge (scientific literature), industry practitioners seeking to protect commercial innovations (patents), and users engaging with the market (social media). Therefore, at a more granular level, considering the incentives that the three audiences reflected in the \ac{dlt}-Corpus have, we investigate two fundamental questions about innovation dynamics in the \ac{dlt}
ecosystem:

\begin{enumerate}
    \item Where do technological concepts first appear, and how do they diffuse across communities?
    \item How does market sentiment relate to research and commercial innovation activity?
\end{enumerate}

These analyses provide insights into the relationship between public discourse, scientific inquiry, and commercial innovation in the \ac{dlt} domain. Most importantly, these analyses serve as introductory demonstrations of use cases for the type of analyses that can be carried out with the \ac{dlt}-Corpus.

\subsection{Technology diffusion across communities} \label{subsec:diffusion_dlt}

We track when key \ac{dlt} concepts first appear within the community of users, academic and industry researchers, or the business-focused community. The \ac{dlt}-Corpus enables this analysis through timestamped documents from scientific literature (\autoref{subsec:industry_and_academy_dataset}), patents (\autoref{subsec:patents_dataset}), and social media (\autoref{subsec:social_media_dataset}).

\paragraph{Technology selection} We focus on three economically significant technologies: (1) \textbf{stablecoins}, given recent regulatory frameworks (US GENIUS Act,\footnote{\url{https://www.congress.gov/bill/119th-congress/senate-bill/394/text}} Europe's MiCAR\footnote{\url{https://www.esma.europa.eu/esmas-activities/digital-finance-and-innovation/markets-crypto-assets-regulation-mica}}) and their role bridging traditional and decentralized finance \cite{HernandezCruz2024NoBank,Liao2022Stablecoins:Banking}. (2) \textbf{\acp{dex}}, holding approximately \$174B in \ac{tvl}\footnote{\url{https://defillama.com/}, accessed 4th April 2025}. (3) \textbf{\acp{amm}}, the mechanism enabling decentralized token swapping \cite{HernandezCruz2025AMM-basedLedger,xu2021dexAmm}.

\paragraph{Findings} Stablecoins (\autoref{fig:StablecoinsPerYear}), \acp{amm} (\autoref{fig:AMMperYear}), and \acp{dex} (\autoref{fig:DEXperYear}) consistently appear first in scientific literature, with researchers maintaining sustained interest over time. This pattern aligns with traditional technology transfer models, in which research precedes commercial application and consumer adoption \cite{Axanova2012U.S.Hypothetical,Amesse2001TechnologyEconomy}.  

\begin{figure*}[t!]
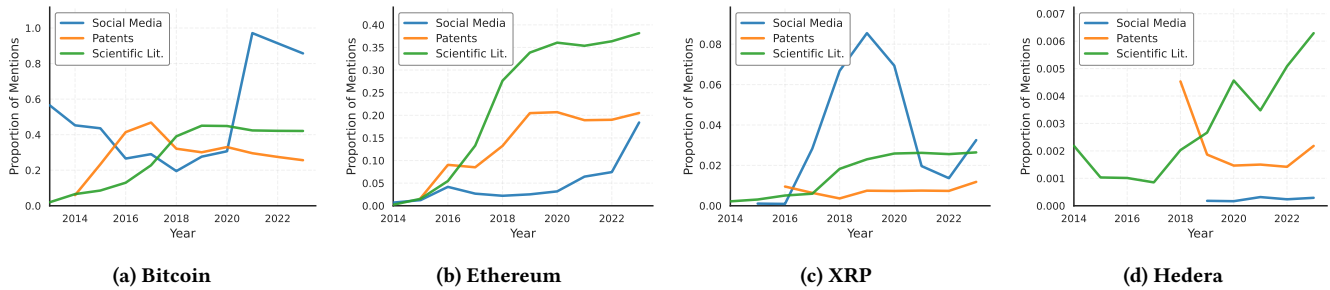

\centering
    \input{figures/bitcoin}
    \hfill
    \input{figures/ethereum}
    \hfill
    \input{figures/xrp}
    \hfill
    \input{figures/hedera}
    \captionsetup{font=small}
    \caption[Proportion of mentions per year for selected cryptocurrencies in the \ac{dlt}-Corpus]{Proportion of mentions per year for selected cryptocurrencies in the \ac{dlt}-Corpus. The y-axis represents the relative frequency of mentions normalized by the total volume of documents in each source per year.}
    \label{fig:CryptosMentionsPerYear}
\end{figure*}

\paragraph{Cryptocurrency mentions vs. technology mentions} We contrast technology diffusion with cryptocurrency mentions to distinguish innovation interest from speculative interest. We analyze Bitcoin, Ethereum, and XRP because they represent the three largest non-stablecoins by market capitalization\footnote{\url{https://coinmarketcap.com/historical/20250302/}\label{fn:coinmarketcap}}). We include Hedera because it uses Hashgraph \cite{Baird2020TheLedgers}, which differs from the blockchain architecture used by most cryptocurrencies. At the time of writing, Hedera is the only non-blockchain \ac{dlt} in the top 20 by market capitalization\footref{fn:coinmarketcap}.

Bitcoin (\autoref{fig:BitcoinMentionsPerYear}) shows high user interest but declining patents and publications that plateau, reflecting the characteristics of a mature consumer asset. Ethereum (\autoref{fig:EthereumMentionsPerYear}) shows increasing publications and patents alongside user interest, reflecting continued innovation in smart contracts and \ac{defi}.\footnote{\url{https://defillama.com/chains}, accessed 4th April 2025} Similar to Ethereum, Hedera (\autoref{fig:HederaMentionsPerYear}) attracts primarily academic interest with limited user engagement, consistent with early-stage technology transfer. XRP (\autoref{fig:XRPMentionsPerYear}) shows sharply declining user engagement around 2020, which coincides with a major lawsuit in the \ac{us} involving this digital asset \cite{Stempel2025SECReuters}. Interestingly, patents and academic and industry research continue growing for XRP before plateauing and slowly picking up again. 

The contrast between cryptocurrency mentions (\autoref{fig:CryptosMentionsPerYear}) and technology mentions (\autoref{fig:TechnologiesMentionsPerYear}) reveals the speculative focus of users on digital assets while researchers and industry practitioners focus on technologies. Then, this raises the question: to what degree does market sentiment affect researchers?

\subsection{Market sentiment and innovation activity} \label{subsec:market-sentiment-analysis}

The \ac{dlt} ecosystem exhibits a unique characteristic: public markets provide real-time feedback on technological developments through cryptocurrency prices and trading activity. This raises an intriguing question about whether market dynamics influence the pace and direction of innovation. Do periods of market enthusiasm correlate with increased research output and patent activity? Or do academic and commercial innovation proceed independently of market sentiment?

\paragraph{Method}
We examine the relationship between social media sentiment and the number of patent filings and scientific publications over time. We classified social media posts according to market sentiment (bullish, bearish, or neutral), using the finetuned LedgerBERT (\autoref{subsec:ledgerBERT}, \autoref{subsec:sentiment_analysis_dataset}), and aggregated them over yearly intervals.

\begin{figure}[t!]

    \centering
    \includegraphics[width=.80\linewidth]{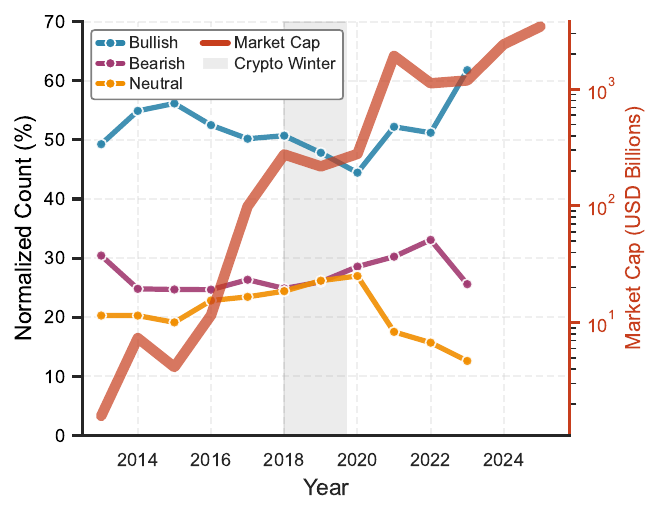}
    \captionsetup{font=small}
    \caption[Market sentiment in social media and yearly growth of global cryptocurrency market capitalization]{Market sentiment in social media and yearly growth of global cryptocurrency market capitalization.}
    \label{fig:MarketSentimentPerYear}
\end{figure}

\paragraph{Findings}
We observe that even during periods of crypto winter (e.g., 2018 to 2019\footnote{https://www.kraken.com/learn/crypto-bull-bear-markets}$^{,}$\footnote{https://finance.yahoo.com/news/contrasting-2022-market-crash-2018-174800300.html}), the user community is overwhelmingly bullish (\autoref{fig:MarketSentimentPerYear}). Additionally, bearish sentiment peaks in 2022, while 2023 sees bullish sentiment grow rapidly as the market recovers. 

Comparing \autoref{fig:DocsTemporalEvolution} with \autoref{fig:MarketSentimentPerYear} reveals that patents and scientific publications follow trajectories largely independent of short-term market sentiment. Instead, innovation activity grows alongside the overall market expansion (\autoref{subsec:diffusion_dlt}).

\section{Discussion}

\paragraph{Divergent community interests}

The user community seems to focus on cryptocurrencies as investments, while researchers appear to concentrate on underlying technologies. New \acfp{dlt} and concepts first appear in the scientific literature before spreading to patents and the user community (\autoref{subsec:diffusion_dlt}), following a traditional technology transfer model \cite{Amesse2001TechnologyEconomy,Axanova2012U.S.Hypothetical}. This pattern suggests that engaging with recently published scientific literature can help identify emerging technologies in the \ac{dlt} field before they become mainstream, potentially creating opportunities for early commercial innovation.

\paragraph{Cryptocurrencies trajectory diverge}
Analysis of specific cryptocurrencies reveals how different factors shape their evolution across communities (\autoref{fig:CryptosMentionsPerYear}). Bitcoin shows declining patent activity and plateauing scientific publications despite sustained interest from the user community, suggesting it is maturing into a consumer-focused digital asset. Ethereum exhibits a different pattern, with growing academic publications and patents alongside user interest, reflecting its continued role in driving innovation through smart contracts and \ac{defi} applications.

XRP shows how external events shape community behavior. For example, XRP's user engagement dropped sharply around 2020 during its legal challenges \cite{Stempel2025SECReuters}, while research activity continued. Hedera primarily attracts academic interest and limited user engagement, suggesting that new \ac{dlt} architectures, such as Hashgraph \cite{Baird2020TheLedgers}, which powers Hedera, can sustain scientific interest without immediate market enthusiasm. However, given that the \ac{dlt}-Corpus indicates the \ac{dlt} field may follow a traditional technology transfer model, Hedera may be in the early stages, with mainstream popularity among users coming later. These patterns show that technological innovation, regulation, and market speculation each independently shape the \ac{dlt} ecosystem.

\paragraph{Research creates economic value through a virtuous cycle}

\autoref{sec:analysis} suggests that research establishes the foundation for the \ac{dlt} field \cite{HernandezCruz2025EvolutionLiterature} that precedes market expansion, while commercial innovation and community discourse respond strongly to market conditions. Then, as cryptocurrency markets grew, increased capital likely funded industry research, leading to more patent filings and heightened community engagement. This creates a virtuous cycle in which foundational research generates innovations that commercial actors and the broader community adopt, develop, and speculate on during periods of market growth, thereby channeling more funding into future research. This pattern benefits the \ac{dlt} field by maintaining a stable research foundation while market-driven activity accelerates technology adoption and deployment.

\section{Conclusions}

We introduce \ac{dlt}-Corpus, a domain resource comprising 2.98 billion tokens from 22.12 million documents across scientific literature (including academic publications and industry whitepapers), \ac{uspto} patents, and social media. Our analysis, serving as an introductory demonstration of the utility of the \ac{dlt}-Corpus, reveals that technologies and concepts typically appear first in scientific literature before appearing in patents and social media, following traditional technology transfer patterns. While social media sentiment remains overwhelmingly bullish, even during crypto winters, scientific and patent activity 
grows less tied to short-term sentiment,
instead tracking overall market expansion.

We release \ac{dlt}-Corpus\footnote{https://huggingface.co/collections/ExponentialScience/dlt-corpus} along with a companion sentiment analysis dataset\footnote{https://huggingface.co/datasets/ExponentialScience/DLT-Sentiment-News} from crowdsourced annotations, the LedgerBERT language model, and the code used to support reproducibility, future research in domain-specific \ac{nlp}, and innovation diffusion analysis for the \ac{dlt} field.

\section{Limitations}

\paragraph{Language coverage} We focus exclusively on English-language data derived from open-access scientific literature, patents, and social media. However, English is the dominant language for web content\footnote{https://www.statista.com/statistics/262946/most-common-languages-on-the-internet/} and nearly all scientific publications \cite{Bahji2023ExclusionPublishers,2023ScientificProblem}.

\paragraph{Domain relevance filtering}
Although we manually revised the filtered scientific literature subset, removing 570 papers (\autoref{subsec:industry_and_academy_dataset}), there is still a possibility that marginally relevant \ac{dlt} papers may remain in the dataset.

\paragraph{Data accessibility and legal compliance trade-offs}
We prioritize data accessibility and legal compliance (see \autoref{sec:ethics}) in constructing the \ac{dlt}-Corpus, which may limit subset sizes and collection from other types of data sources, like news (\autoref{subsec:sentiment-evaluation}), but reduces legal barriers for academic and commercial use of the \ac{dlt}-Corpus. This emphasis addresses growing industry concerns about regulatory and copyright risks in \ac{ai} development. S\&P 500 companies have acknowledged in hundreds of corporate filings and executive transcripts that legal and regulatory risks are primary concerns in their \ac{ai} adoption \cite{Heikkila2025AmericasUpsides}, which is likely amplified by ongoing copyright lawsuits against OpenAI \cite{Brittain2025JudgeReuters,Pope2024NYTReview}, Microsoft \cite{Pope2024NYTReview}, Anthropic \cite{Jamali2025AINews}, and Meta \cite{Knibbs2025MetaWIRED} over alleged copyright violations from data collected and used for training their \acp{llm}.

\section{Ethics} \label{sec:ethics}

The sentiment analysis dataset and \ac{dlt}-Corpus could enable market manipulation or coordinated trading strategies. We acknowledge this risk but note that such tools are already widely available, and our contribution primarily advances research transparency.

All data in \ac{dlt}-Corpus derives from publicly available sources. For the scientific literature subset, we exclusively use open-access publications with documented licensing (via Semantic Scholar\footnote{https://www.semanticscholar.org/}) to ensure redistribution rights. Patent data is collected from \ac{uspto}, which explicitly states in its Terms of Use that patent text is 
\blockquote{typically not subject to copyright restrictions}\footnote{https://www.uspto.gov/terms-use-uspto-websites}. The social media subset aggregates previously published academic and industry datasets (see \autoref{subsec:social_media_dataset}) collected before Twitter/X implemented significant API access restrictions and pricing changes in 2023 \cite{Davidson2023Platform-controlledScience}. By posting publicly, users granted Twitter licenses to make their content \blockquote{available to other companies, organizations or individuals} for distribution based on Twitter/X's \ac{tcs} at the time.\footnote{https://x.com/en/tos/previous/version\_17}$^{,}$\footnote{https://x.com/en/tos/previous/version\_18}

Additionally, our research complies with \ac{gdpr} principles, particularly Article 89\footnote{https://gdpr.eu/article-89-processing-for-archiving-purposes-scientific-or-historical-research-purposes-or-statistical-purposes/} academic research exemptions for processing publicly available data.\footnote{https://gdpr-text.com/read/article-85/} Similarly, following \ac{gdpr} guidelines (Article 5(1)(c)\footnote{https://gdpr.eu/article-5-how-to-process-personal-data/}), we apply data minimization\footnote{https://europa.eu/youreurope/business/dealing-with-customers/data-protection/data-protection-gdpr/index\_en.htm} by excluding usernames from social media posts, retaining only text content and timestamps. While Twitter's \ac{tcs} permitted the collection of public content, including username, we recognize that users' privacy expectations may have changed since the original posting, and username removal reduces the risk of cross-platform tracking or other potential harms without compromising the utility of our dataset for the research purposes outlined in this work.

\section{Acknowledgements}
We thank Max Bartolo for his valuable and constructive feedback on our dataset evaluation and language model training. The project is partially supported by UK Centre for Blockchain Technologies through Ripple's University Blockchain Research Initiative (UBRI) \cite{Feng2022}.

\bibliographystyle{ACM-Reference-Format}
\bibliography{references}

\newpage
\appendix

\section{Additional datasets documentation}
\label{sec:additional-datasets-documentation}

\subsection{Additional \acs{dlt}-Corpus documentation} \label{sec:additional-dlt-corpus-documentation}

\subsection*{Uses} \label{subsec:use-dlt-corpus}

\textbf{Intended use:} \ac{nlp} research, language model development, innovation studies, text mining in \ac{dlt} domain, and other social and computational linguistic studies.

\textbf{Unsuitable uses:} Identification of specific individuals, creating investment advice systems without proper disclaimers, and applications requiring post-2023 social media data.

\textbf{Impact:} May enable market manipulation if misused. Researchers should implement appropriate safeguards.

\subsection*{Distribution}

\textbf{Access:} \url{https://huggingface.co/collections/ExponentialScience/dlt-corpus}

\textbf{License:} 

\textit{Scientific literature:} Mixed open-access licenses (CC-BY, CC-BY-SA, CC0, and other permissive licenses). Individual license information is included in metadata where available.
\textit{Patents:} Public domain under \ac{uspto}'s \ac{tcs}. Patent text is typically not subject to copyright restrictions per \ac{uspto}'s \ac{tcs}\footnote{\url{https://www.uspto.gov/terms-use-uspto-websites}}.
\textit{Social media:} Released under CC-BY-NC 4.0 for research purposes. Collected before changes in Twitter / X's \ac{tcs} in 2023\footnote{\url{https://x.com/en/tos/previous/version\_18}}$^{,}$\footnote{\url{https://x.com/en/tos/previous/version\_17}}, permitting academic research use \cite{Davidson2023Platform-controlledScience}.

\subsection*{Maintenance}

\textbf{Updates:} Currently static snapshot. Future versions may expand scientific literature and patents, but would likely not include post-2023 social media.

\subsection{Sentiment analysis dataset} \label{subsec:sentiment-analysis-documentation}

\subsection*{Motivation}

\textbf{Purpose:} The DLT Sentiment Analysis Dataset\footnote{\url{https://huggingface.co/datasets/ExponentialScience/DLT-Sentiment-News}} was created to support sentiment analysis research in the \ac{dlt} domain, addressing the lack of high-quality labeled data that captures domain-specific sentiment expressed by cryptocurrency community members.

\textbf{Creators:} Walter Hernandez Cruz, Peter Devine, Nikhil Vadgama, Paolo Tasca, Jiahua Xu

\subsection*{Composition}

\textbf{Content:} 23,301 examples with 1.85M tokens (average 79.51 tokens per example). 

\textbf{Labels:} Three sentiment dimensions with three categories each. \textit{Market direction}: bullish, bearish, neutral. \textit{Content characteristics}: important, lol, neutral. \textit{Engagement quality}: liked, disliked, neutral.

\textbf{Temporal coverage:} January 2021 to May 2025

\textbf{Language:} English

\textbf{Missing data:} None.

\textbf{Confidentiality:} No private or confidential data. All content derived from publicly available cryptocurrency news articles headlines (and brief descriptions) voted on by CryptoPanic users.

\textbf{Data fields}: \autoref{tab:fields-sentiment} describes the fields in the sentiment analysis dataset.

\begin{table}[t]

\footnotesize
\centering
\captionsetup{size=footnotesize}
\caption{Fields in Sentiment Analysis dataset}
\label{tab:fields-sentiment}

\begin{tabular}{ll}
\toprule
\textbf{Field} & \textbf{Description} \\
\midrule
\texttt{title} & Article headline \\
\texttt{description} & Article summary \\
\texttt{text} & Combined title and description \\
\texttt{timestamp} & Publication time \\
\texttt{url} & Original article link \\
\texttt{source\_url} & CryptoPanic news link \\
\texttt{market\_direction} & Bullish/bearish/neutral label \\
\texttt{engagement\_quality} & Liked/disliked/neutral label \\
\texttt{content\_characteristics} & Important/lol/neutral label \\
\texttt{vote\_counts} & Votes per category \\
\texttt{total\_votes} & Total vote count \\
\texttt{total\_tokens} & Token count \\
\bottomrule
\end{tabular}
\end{table}

\subsection*{Collection}

\textbf{Source:} CryptoPanic platform, where cryptocurrency community users vote on news articles' headlines and their brief descriptions across multiple sentiment categories.

\textbf{Annotation method:} Crowdsourced voting by active cryptocurrency users, providing domain expertise. Vote percentages normalized by total engagement, filtered using median minimum votes, with 25th and 75th percentiles as classification boundaries (\autoref{subsec:sentiment_analysis_dataset}).

\subsection*{Preprocessing}

\textbf{Label assignment:} Percentile-based classification to mitigate popularity bias. Articles below the 25th percentile are labeled negative, above the 75th percentile are labeled positive, and those between are labeled neutral for each dimension (see \autoref{subsec:sentiment_analysis_dataset} for more details).

\textbf{Quality control:} Minimum vote threshold applied to exclude articles with insufficient community engagement (\autoref{subsec:sentiment_analysis_dataset}).

\subsection*{Uses}

\textbf{Intended use:} Sentiment analysis research, domain-specific model evaluation, market sentiment studies in \ac{dlt} domain.

\textbf{Unsuitable uses:} Investment decision systems without proper disclaimers, identifying individual voters, and applications requiring real-time sentiment.

\textbf{Impact:} May enable market manipulation if misused. Researchers should implement appropriate safeguards and ethical guidelines.

\subsection*{Distribution}

\textbf{Access:} \url{https://huggingface.co/datasets/ExponentialScience/DLT-Sentiment-News}

\textbf{License:} CC-BY-NC 4.0 for research purposes. Derived from publicly available CryptoPanic data with crowdsourced community annotations. Data collected via CryptoPanic's free API between March and May 2025. To the best of our knowledge, the \ac{tcs} at the time of collection (cryptopanic.com/terms/) contained no restrictions on academic research use or redistribution.

\subsection*{Maintenance}

\textbf{Updates:} Currently static snapshot. Future versions may expand temporal coverage or add additional sentiment dimensions.

\end{document}